\documentclass[11pt,a4paper]{article}

\usepackage[utf8]{inputenc}
\usepackage[T1]{fontenc}
\usepackage{amsmath,amssymb}
\usepackage{graphicx}
\usepackage{booktabs}
\usepackage{multirow}
\usepackage{hyperref}
\usepackage{xcolor}
\usepackage{geometry}
\usepackage{caption}
\usepackage{subcaption}
\usepackage{pgfplots}
\usepackage{pgfplotstable}
\usepackage{array}
\usepackage{tabularx}
\usepackage{float}
\usepackage{enumitem}
\usepackage{fancyhdr}
\usepackage{titlesec}

\pgfplotsset{compat=1.18}

\hypersetup{
    colorlinks=true,
    linkcolor=blue!70!black,
    citecolor=blue!70!black,
    urlcolor=blue!70!black
}

\title{%
\vspace{-1cm}
\textbf{\LARGE Pushing the Limits of On-Device Streaming ASR:\\
A Compact, High-Accuracy English Model\\
for Low-Latency Inference}\\[0.3cm]

}

\author{%
Nenad Banfic, David Fan, Kunal Vaishnavi, Sam Kemp,\\[6pt]
Sunghoon Choi, Rui Ren, Sayan Shaw, Meng Tang\\[8pt]
{\small CoreAI, Microsoft}
}

\date{April 2026}

\begin{document}

\maketitle
\thispagestyle{fancy}

% ============================================================
\begin{abstract}
Deploying high-quality automatic speech recognition (ASR) on edge devices requires models that jointly optimize accuracy, latency, and memory footprint while operating entirely on CPU without GPU acceleration. We conduct a systematic empirical study of state-of-the-art ASR architectures, encompassing encoder--decoder, transducer, and LLM-based paradigms, evaluated across batch, chunked, and streaming inference modes. Through a comprehensive benchmark of over 50 configurations spanning OpenAI Whisper, NVIDIA Nemotron, Parakeet TDT, Canary, Conformer Transducer, and Qwen3-ASR, we identify NVIDIA's Nemotron Speech Streaming as the strongest candidate for real-time English streaming on resource-constrained hardware. We then re-implement the complete streaming inference pipeline in ONNX Runtime and conduct a controlled evaluation of multiple post-training quantization strategies, including importance-weighted k-quant, mixed-precision schemes, and round-to-nearest quantization, combined with graph-level operator fusion. These optimizations reduce the model from 2.47\,GB to as little as 0.67\,GB while maintaining word error rate (WER) within 1\% absolute of the full-precision PyTorch baseline. Our recommended configuration, the int4 k-quant variant, achieves 8.20\% average streaming WER across eight standard benchmarks, running comfortably faster than real-time on CPU with 0.56\,s algorithmic latency, establishing a new quality–efficiency Pareto point for on-device streaming ASR.
\end{abstract}

\vspace{0.5cm}
\hrule
\vspace{0.5cm}

% ============================================================
\section{Introduction}
\label{sec:intro}

In this report, we present a streaming ASR system optimized for on-device deployment via ONNX Runtime. Built on NVIDIA's Nemotron Speech and optimized through quantization and operator fusion, it achieves competitive results on standard English benchmarks while fitting under 1\,GB and running faster than real-time on CPU with sub-second latency.

Automatic Speech Recognition has undergone rapid improvements in recent years, with large-scale models such as OpenAI Whisper~\cite{whisper}, NVIDIA NeMo models~\cite{nemo}, and Qwen3-ASR~\cite{qwen3asr} pushing the boundaries of accuracy on standard English benchmarks. However, the best-performing models---Qwen3-ASR-1.7B at 5.90\% WER, Parakeet TDT-0.6B-v3 at 6.32\%, and Canary-1B-v2 at 7.15\%---are all primarily batch-oriented architectures requiring 2--7\,GB of memory and GPU inference. Deploying high-quality ASR in edge scenarios, where compute is limited to CPUs, latency must be minimal, and memory budgets are tight, remains a significant challenge. Prior work has studied strong ASR architectures, streaming inference strategies, and low-bit quantization, but these directions are often evaluated in isolation. This report focuses on their intersection: identifying a strong streaming architecture under a controlled public benchmark suite and compressing it for practical edge deployment.

Our primary objective was to identify a high-quality English streaming ASR model that operates under the lowest possible resource utilization---minimizing model size, memory consumption, CPU load, and algorithmic latency without sacrificing transcription quality---enabling speech recognition directly on user devices without reliance on cloud infrastructure. This translates to four concrete constraints:

\begin{enumerate}[leftmargin=2cm]
    \item \textbf{Streaming capability:} The model must produce transcriptions with sub-second latency, processing audio in small chunks rather than requiring the full utterance.
    \item \textbf{High accuracy:} Word error rates must be competitive across diverse English domains (meetings, earnings calls, broadcast, read speech, spontaneous speech).
    \item \textbf{Minimal resource utilization:} The model must fit within the memory and storage constraints of consumer hardware, ideally under 1\,GB, while running comfortably faster than real-time on CPU alone.
    \item \textbf{CPU-only inference:} The model must not require a GPU, enabling deployment on the widest range of edge hardware.
\end{enumerate}

To arrive at this model, we conducted a comprehensive evaluation of six model families across eight English benchmark datasets, testing over 50 distinct configurations of architecture, precision, chunking strategy, and quantization level. This report documents our methodology, comparative findings, and the optimization pipeline that produced the final on-device streaming ASR model.

% ============================================================
\section{Evaluated Model Families}
\label{sec:models}

We evaluated the following model families, covering the major paradigms in modern ASR:

\begin{table}[H]
\centering
\caption{Overview of evaluated model families.}
\label{tab:model_families}
\begin{tabular}{@{}llcc@{}}
\toprule
\textbf{Model} & \textbf{Architecture} & \textbf{Modes} & \textbf{Size} \\
\midrule
Whisper Large-v3-Turbo & Encoder--Decoder & Batch & 1.62\,GB \\
Whisper Small & Encoder--Decoder & Chunk/Batch & 0.97\,GB \\
Nemotron-0.6B & Cache-aware Transducer & Stream/Batch & 2.47\,GB \\
Parakeet TDT-0.6B-v3 & TDT Transducer & Chunk/Batch & 2.51\,GB \\
Canary-1B-v2 & AED + AlignAtt & Chunk/Batch & 6.36\,GB \\
Conformer Trans. XL & Conformer Transducer & Chunk & 2.58\,GB \\
Qwen3-ASR-1.7B & LLM-based ASR & Batch/Chunk & 4.70\,GB \\
Qwen3-ASR-0.6B & LLM-based ASR & Batch/Chunk & 1.88\,GB \\
\bottomrule
\end{tabular}
\end{table}

\subsection{Whisper Models}
OpenAI's Whisper models use an encoder--decoder architecture with cross-attention. While highly accurate in batch mode, they are not natively designed for streaming. We also evaluated an ORT CUDA FP16 variant optimized with Olive~\cite{olive}, and chunked Whisper Small inference via Faster-Whisper~\cite{fasterwhisper} using a sliding-window approach with overlapping segments.

\subsection{Parakeet TDT and Conformer Transducer}
We group these two models together as both are transducer-based architectures. NVIDIA's Parakeet TDT-0.6B-v3~\cite{parakeet} uses a Token-and-Duration Transducer architecture with multilingual support, making it a strong candidate for future language expansion if chunked streaming proves viable. We tested it extensively across 15+ chunking configurations to characterize the relationship between chunk size, context length, and WER on English benchmarks. The Conformer Transducer XL~\cite{conformerxl} is a larger English-only conformer transducer model also evaluated in chunked mode.

\subsection{Qwen3-ASR}
Qwen3-ASR represents the emerging LLM-based approach to ASR, where a language model backbone is adapted for speech. We evaluated both the 1.7B and 0.6B variants in batch and chunked modes with varying stride lengths on English benchmarks.

\subsection{NVIDIA Nemotron Speech Streaming}
NVIDIA's Nemotron Speech Streaming~\cite{nemotron} is a cache-aware streaming conformer transducer with approximately 600 million parameters. Unlike the batch-only models above, it is purpose-built for real-time streaming: the encoder uses a chunked attention mechanism that caches prior context across chunks, enabling flexible latency--accuracy trade-offs by adjusting the streaming configuration at inference time without retraining.

% ============================================================
\section{Evaluation Methodology}
\label{sec:methodology}

\subsection{Benchmark Datasets}
All models were evaluated on eight standard English ASR benchmarks from the ESB (End-to-end Speech Benchmark) suite, spanning diverse acoustic conditions and speaking styles. For batch-mode evaluation, we use the Open ASR Leaderboard~\cite{open_asr} framework. To support streaming and chunked evaluation, we extended this framework to measure streaming WER by feeding audio in chunks and assembling the final transcript from chunk-level outputs. Our extended evaluation code is publicly available.\footnote{\url{https://github.com/nenad1002/open_asr_leaderboard}}

\begin{table}[H]
\centering
\caption{Evaluation benchmark datasets.}
\label{tab:datasets}
\begin{tabular}{@{}lll@{}}
\toprule
\textbf{Dataset} & \textbf{Domain} & \textbf{Characteristics} \\
\midrule
AMI & Meeting transcription & Overlapping speech, far-field \\
Earnings22 & Financial earnings calls & Domain-specific terminology \\
GigaSpeech & Internet audio & Diverse topics and acoustics \\
LibriSpeech Clean & Audiobook (clean) & Read speech, studio quality \\
LibriSpeech Other & Audiobook (other) & Read speech, noisier conditions \\
SPGISpeech & Financial transcription & Professional dictation \\
TED-LIUM & TED talks & Prepared speech, varied topics \\
VoxPopuli & European Parliament & Spontaneous speech, accented \\
\bottomrule
\end{tabular}
\end{table}

\subsection{Metrics}
\begin{itemize}[leftmargin=1.5cm]
    \item \textbf{Word Error Rate (WER):} Standard metric computed per dataset. We report individual WERs and the unweighted average across all eight datasets. In batch mode, the full utterance is processed at once and the WER is computed on the final output. In streaming and chunked modes, we report the \emph{streaming WER}: the transcription is assembled from all chunk outputs after the entire audio has been processed, and WER is computed on the concatenated result.
    \item \textbf{Real-Time Factor (RTFx):} Defined as:
    \[
    \text{RTFx} = \frac{\text{audio duration}}{\text{wall-clock processing time}}
    \]
    where processing time includes all stages of the inference pipeline (audio preprocessing, mel extraction, encoder, decoder, and joiner inference). An RTFx of 5$\times$ means the model processes audio five times faster than real-time. RTFx is hardware-dependent; we report it at \texttt{batch\_size=1} as the utterance-level average, with the specific hardware noted in the section below.
    For streaming deployment, the critical requirement is that \emph{per-chunk} RTFx consistently exceeds 1.0$\times$ to avoid audio dropout. Since absolute RTFx values are hardware-dependent, RTFx should be compared relative to the baseline configuration on the same hardware rather than interpreted in isolation.
    \item \textbf{Latency (Delay):} The algorithmic delay introduced by the chunking/streaming configuration, determined by the chunk size and right context. For streaming ASR, we can also define the effective latency, bounded by the algorithmic delay plus the compute time for a single chunk: $\text{effective latency} \approx \text{delay} + \frac{\text{chunk duration}}{\text{RTFx}}$. When the algorithmic delay equals the chunk duration (as in all Nemotron configurations we evaluate), this simplifies to $\text{effective latency} \approx \text{delay} \cdot \left(1 + \frac{1}{\text{RTFx}}\right)$.
    \item \textbf{Model Size:} On-disk size of the model weights.
    \item \textbf{Batch-to-Stream Factor (BSF):} The ratio of streaming WER to batch WER for the same model:
    \[
    \text{BSF} = \frac{\text{Streaming WER}}{\text{Batch WER}}
    \]
    A BSF of 1.0 indicates no accuracy loss from streaming; values above 1.0 quantify how much a model degrades when moving from batch to real-time operation.
\end{itemize}

\subsection{Hardware}
All CPU measurements use an AMD EPYC 7V12 64-Core Processor (AVX2, 2.45\,GHz), with inference pinned to 32 cores to ensure consistent and reproducible results across runs. GPU measurements use an NVIDIA H100 (CUDA). The specific hardware and batch size are noted in each table caption in the appendix, where appropriate.

% ============================================================
\section{Batch-Mode Comparison}
\label{sec:batch}

We first evaluated all models in batch (offline) mode to establish accuracy upper bounds. All results reported in this section reflect our own evaluation under the controlled setup described in Section~\ref{sec:methodology}; reported values may differ from officially published numbers due to differences in hardware, text normalization, or evaluation methodology. Table~\ref{tab:batch_results} shows the results.

\begin{table}[H]
\centering
\caption{Batch-mode WER (\%) comparison across models.}
\label{tab:batch_results}
\resizebox{\textwidth}{!}{%
\begin{tabular}{@{}lccccccccc@{}}
\toprule
\textbf{Model} & \textbf{Avg} & \textbf{AMI} & \textbf{Earn.} & \textbf{Giga} & \textbf{LS-c} & \textbf{LS-o} & \textbf{SPGI} & \textbf{TED} & \textbf{VoxP.} \\
\midrule
Qwen3-ASR-1.7B & \textbf{5.90} & 11.76 & \textbf{10.26} & \textbf{8.75} & \textbf{1.60} & \textbf{3.41} & 2.83 & \textbf{2.28} & 6.34 \\
Qwen3-ASR-0.6B & 6.69 & 13.77 & 11.03 & 9.16 & 2.12 & 4.47 & 3.04 & 2.83 & 7.09 \\
Parakeet TDT-0.6B-v3 & 6.32 & 11.39 & 11.19 & 9.57 & 1.92 & 3.59 & 3.98 & 2.80 & \textbf{6.09} \\
Canary-1B-v2 & 7.15 & 16.01 & 11.79 & 10.82 & 2.18 & 3.56 & \textbf{2.28} & 4.29 & 6.25 \\
Nemotron-0.6B & 7.07 & \textbf{11.16} & 12.38 & 11.35 & 2.28 & 4.83 & 2.63 & 4.42 & 7.48 \\
Whisper-v3-Turbo & 7.83 & 16.13 & 11.63 & 10.14 & 2.10 & 4.24 & 2.97 & 3.57 & 11.87 \\
Whisper-v3-Turbo (ORT) & 7.52 & 16.36 & 11.38 & 10.13 & 2.17 & 4.24 & 2.93 & 3.62 & 9.30 \\
Whisper Small.en & 8.59 & 17.93 & 12.97 & 11.35 & 3.05 & 7.25 & 3.60 & 4.07 & 8.50 \\
\bottomrule
\end{tabular}%
}
\vspace{2pt}
\par\noindent{\scriptsize Earn.\,=\,Earnings22, Giga\,=\,GigaSpeech, LS-c/o\,=\,LibriSpeech Clean/Other, SPGI\,=\,SPGISpeech, TED\,=\,TED-LIUM, VoxP.\,=\,VoxPopuli.}
\end{table}

Qwen3-ASR-1.7B achieves the best batch WER among the evaluated models, however, its size of 4.70\,GB exceeds our size requirements by far. Qwen3-ASR-0.6B, Parakeet TDT-0.6B-v3, and Nemotron-0.6B are all promising, so we next evaluate them in streaming and chunked settings.

% ============================================================
\section{Streaming and Chunked Mode Analysis}
\label{sec:streaming}

\subsection{Parakeet TDT Chunking Analysis}
\label{sec:parakeet}

\begin{figure}[H]
\centering
\includegraphics[width=0.85\textwidth]{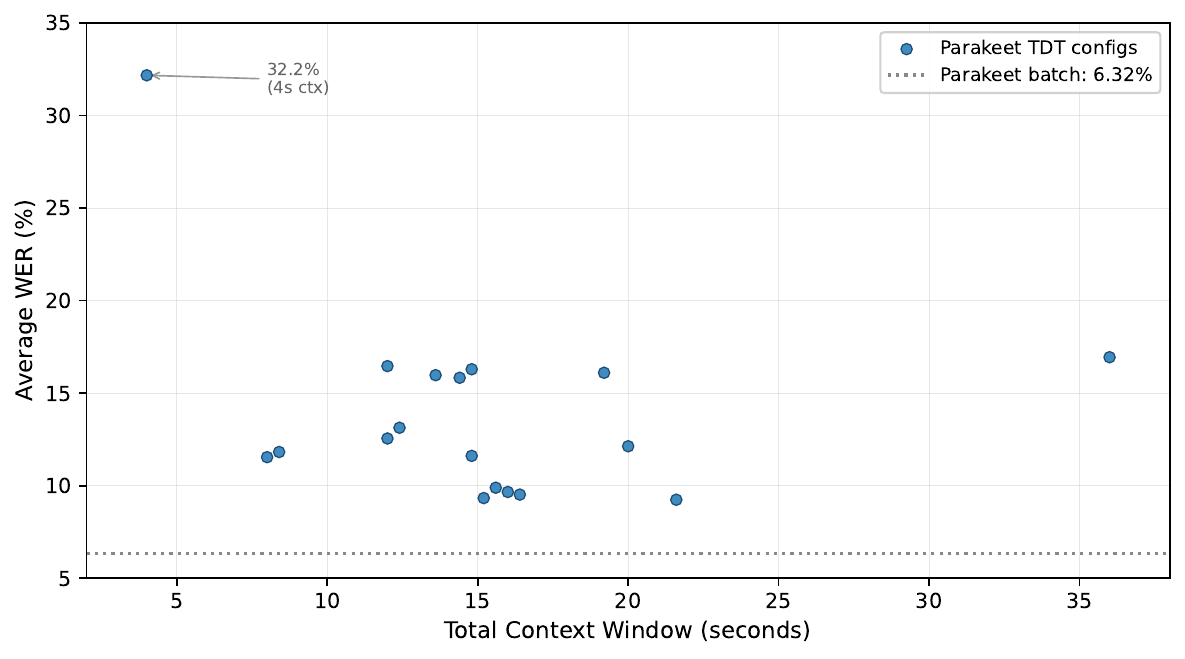}
\caption{Parakeet TDT-0.6B-v3: WER vs.\ total context window across 18 chunking configurations. Even the best chunked configuration (9.22\%) degrades substantially from the 6.32\% batch baseline.}
\label{fig:parakeet_chunks}
\end{figure}

We first investigated whether batch-oriented transducer models can be adapted to chunked streaming. We conducted an extensive search over chunking configurations for Parakeet TDT-0.6B-v3, testing various total context and history lengths while keeping the chunk size (delay) at 2.4\,s, to understand the sensitivity of transducer models to chunk boundaries. Figure~\ref{fig:parakeet_chunks} visualizes the results.

The best Parakeet chunked configuration achieves 9.22\% WER, a 46\% relative increase over its 6.32\% batch WER. This analysis confirms that models not specifically designed for streaming incur substantial penalties when adapted to chunked operation, motivating our focus on natively streaming architectures.

\subsection{Nemotron Streaming Configurations}

The Nemotron streaming configuration is specified as \texttt{(chunk\_size, left\_context, shift\_size)} in units of 80\,ms. For example, the configuration \texttt{(7, 10, 7)} corresponds to a 560\,ms chunk, 800\,ms of cached history per chunk, and a 5.6\,s effective history window. A key advantage of this architecture is the ability to tune latency--accuracy trade-offs by adjusting these parameters at inference time without retraining. Table~\ref{tab:nemotron_streaming} summarizes results across several configurations.

\begin{table}[H]
\centering
\caption{Nemotron-0.6B streaming configurations.}
\label{tab:nemotron_streaming}
\resizebox{\textwidth}{!}{%
\begin{tabular}{@{}lccccccccccc@{}}
\toprule
\textbf{Config} & \textbf{Delay} & \textbf{History} & \textbf{Avg} & \textbf{AMI} & \textbf{Earn.} & \textbf{Giga} & \textbf{LS-c} & \textbf{LS-o} & \textbf{SPGI} & \textbf{TED} & \textbf{VoxP.} \\
\midrule
Batch (offline) & -- & -- & 7.07 & 11.16 & 12.38 & 11.35 & 2.28 & 4.83 & 2.63 & 4.42 & 7.48 \\
(70, 2, 70) & 5.6\,s & 11.2\,s & 7.10 & 11.31 & 12.46 & 11.39 & 2.28 & 4.84 & 2.64 & 4.42 & 7.47 \\
(14, 2, 14) & 1.12\,s & 2.24\,s & 7.46 & 11.38 & 12.49 & 11.39 & 2.41 & 5.07 & 2.71 & 4.50 & 9.75 \\
(7, 10, 7) & 0.56\,s & 5.6\,s & 7.28 & 11.80 & 12.64 & 11.50 & 2.33 & 5.08 & 2.72 & 4.46 & 7.69 \\
(7, 2, 7) & 0.56\,s & 1.12\,s & 8.51 & 11.93 & 13.93 & 11.76 & 3.43 & 6.00 & 3.47 & 5.11 & 12.49 \\
(2, 35, 2) & 0.16\,s & 5.6\,s & 8.04 & 14.82 & 13.66 & 12.44 & 2.49 & 5.52 & 2.90 & 4.72 & 7.79 \\
(1, 70, 1) & 0.08\,s & 5.6\,s & 8.91 & 17.76 & 14.64 & 13.33 & 2.61 & 5.93 & 3.73 & 5.06 & 8.24 \\
\bottomrule
\end{tabular}%
}
\end{table}

\begin{figure}[H]
\centering
\includegraphics[width=0.85\textwidth]{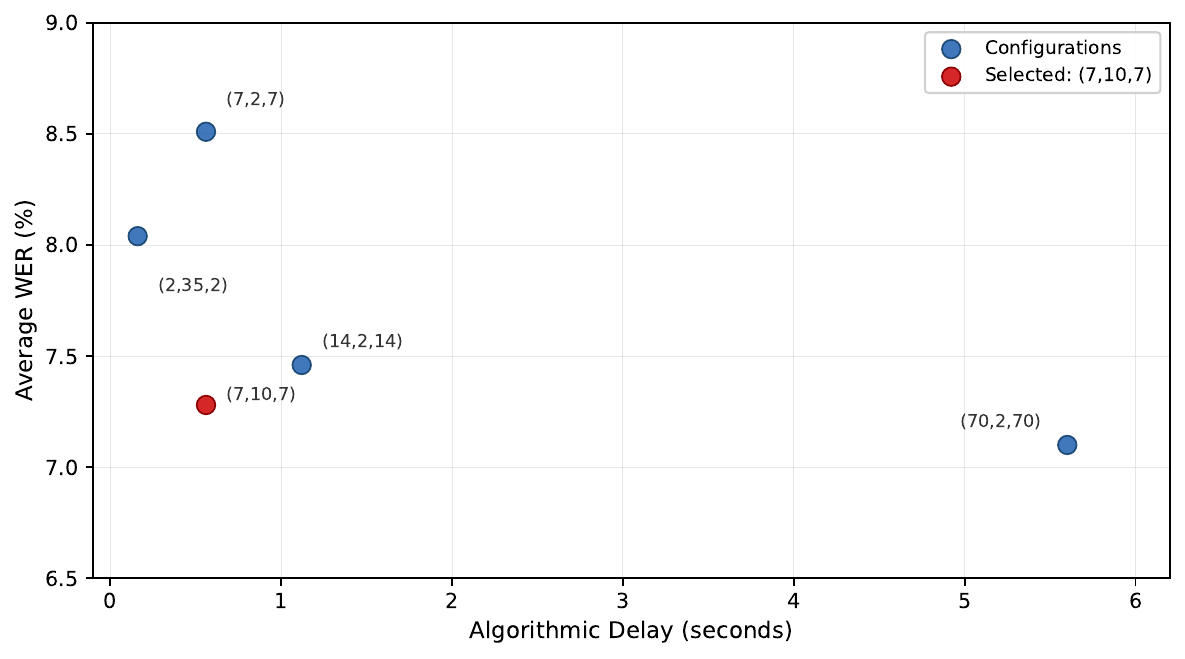}
\caption{Nemotron-0.6B: Delay vs.\ WER trade-off across streaming configurations. The configuration (7,10,7) with 0.56\,s delay and 5.6\,s history achieves the best balance, reaching 7.28\% WER, only 0.21\% above the batch baseline.}
\label{fig:delay_wer}
\end{figure}

The configuration \texttt{(7, 10, 7)} emerges as the optimal operating point: it provides only 0.56\,s of algorithmic delay while achieving 7.28\% average WER, merely 0.21\% absolute above the offline batch baseline. In contrast, the best Parakeet chunked result (9.22\%) is 1.94 percentage points worse at $4\times$ higher latency. The key insight is that sufficient \emph{history context} (5.6\,s via 10 left chunks) is critical. It's interesting to compare (7, 10, 7) at 7.28\% WER with (7, 2, 7) at 8.51\% WER, where the only difference is reduced history.

\subsection{Comparison with Other Streaming/Chunked Models}

To validate that Nemotron is indeed the strongest streaming candidate, we compared the best streaming or low-latency configuration from each model family using the Batch-to-Stream Factor (BSF) defined in Section~\ref{sec:methodology}. Table~\ref{tab:streaming_comparison} reports the results.

\begin{table}[H]
\centering
\caption{Cross-model streaming comparison (batch\_size=1, CPU).}
\label{tab:streaming_comparison}
\resizebox{\textwidth}{!}{%
\begin{tabular}{@{}lccccccccccccc@{}}
\toprule
\textbf{Model} & \textbf{Size} & \textbf{Delay} & \textbf{Avg} & \textbf{BSF} & \textbf{RTFx} & \textbf{AMI} & \textbf{Earn.} & \textbf{Giga} & \textbf{LS-c} & \textbf{LS-o} & \textbf{SPGI} & \textbf{TED} & \textbf{VoxP.} \\
\midrule
Nemotron-0.6B & 2.47\,GB & 0.56\,s & 7.28 & \textbf{1.03} & 2.46 & 11.80 & 12.64 & 11.50 & 2.33 & 5.08 & 2.72 & 4.46 & 7.69 \\
Qwen3-ASR-1.7B & 4.70\,GB & 2.4\,s & 10.45 & 1.77 & 0.49 & 16.97 & 16.92 & 12.29 & 4.95 & 7.69 & 6.70 & 6.34 & 11.75 \\
Parakeet TDT-0.6B-v3 & 2.51\,GB & 2.4\,s & 12.83 & 2.03 & 1.38 & 16.73 & 19.81 & 15.27 & 6.63 & 8.25 & 17.82 & 9.95 & 8.19 \\
Conformer Trans. XL & 2.58\,GB & 2.4\,s & 11.06 & -- & 1.27 & 22.57 & 24.44 & 14.05 & 2.11 & 3.83 & 7.95 & 5.36 & 8.18 \\
Canary-1B-v2 & 6.36\,GB & 4.8\,s & 12.45 & 1.74 & 2.93 & 22.10 & 22.04 & 14.60 & 4.87 & 5.96 & 4.42 & 7.64 & 17.98 \\
\bottomrule
\end{tabular}%
}
\end{table}

Key observations:
\begin{itemize}
    \item \textbf{Nemotron-0.6B achieves the best streaming WER} (7.28\%) among all models tested, while also having the \emph{lowest} latency at 0.56\,s and a BSF of \textbf{1.03}, essentially no degradation from batch mode.
    \item Qwen3-ASR-1.7B, despite being the best batch model among the tested models, degrades significantly in chunked mode (BSF = 1.77) and runs below real-time on CPU (RTFx = 0.49), making it unsuitable for edge deployment. The Qwen3-ASR-0.6B model showed similar results, as shown in the appendix.
    \item Parakeet TDT and Canary show substantial streaming degradation (BSF $\geq$ 1.74), with WER almost doubling despite operating at $4\times$ higher latency than Nemotron.
    \item While Parakeet TDT achieves higher RTFx than Nemotron on GPU (see Table~\ref{tab:app_parakeet} in the appendix), it is notably slower on CPU (RTFx 1.38 vs.\ 2.46). We hypothesize that Parakeet's larger chunks benefit from GPU parallelism over wide tensors, whereas on CPU, where parallelism is limited, Nemotron's smaller, cache-efficient chunks incur less per-step compute and better exploit the CPU memory hierarchy.
\end{itemize}

% ============================================================
\section{ONNX Runtime Integration and Model Optimization}
\label{sec:engineering}

Having identified Nemotron-0.6B with the \texttt{(7, 10, 7)} configuration as our base model, we built a complete end-to-end streaming inference pipeline and applied ONNX Runtime quantization to optimize it for CPU-only edge deployment.

\subsection{Streaming Inference Architecture}

The original Nemotron model relies on NVIDIA's NeMo framework and PyTorch for inference---neither of which is suitable for lightweight edge deployment. To enable efficient CPU inference, we re-implemented the full streaming ASR pipeline natively in ONNX Runtime~\cite{onnxruntime} for efficient cross-platform deployment.

\medskip
The key inference-level design decisions were:

\begin{enumerate}
    \item \textbf{Three-graph decomposition.} Rather than exporting a single monolithic ONNX graph, we decompose the model into three independently optimizable ONNX sessions: the encoder (cache-aware FastConformer), the decoder (LSTM prediction network), and the joiner. This enables per-component quantization---for example, quantizing the encoder more aggressively than the decoder---and allows ONNX Runtime to apply graph-level optimizations independently to each component. In particular, fusing multi-head attention into a single kernel yielded significant speedups for the encoder, which dominates inference time.
    
    \item \textbf{Stateful streaming with zero-copy cache management.} The encoder maintains rolling cache tensors across chunks for channel and temporal context, while the decoder preserves LSTM hidden and cell states between decoding steps. We designed the inference loop to update these caches in-place between chunks, avoiding redundant memory allocations and copies that would otherwise dominate CPU inference latency for short audio segments.
    
    \item \textbf{Native mel spectrogram extraction.} Rather than depending on Python-based audio preprocessing, we implemented NeMo-compatible log-mel feature extraction directly in the inference runtime. This includes a ring-buffer pre-encode cache that carries overlapping mel frames between chunks, ensuring acoustic continuity at chunk boundaries without recomputing features.
    
    \item \textbf{RNNT greedy decoding.} The inference loop implements RNNT greedy decoding as a state machine that iterates over encoder time steps, querying the joiner at each step until a blank token is emitted or a per-step symbol limit is reached. This avoids the overhead of beam search while maintaining accuracy for streaming scenarios.
\end{enumerate}

\subsection{Quantization Methods}
\label{sec:quant_methods}

Post-training quantization methods can be broadly divided into calibration-based approaches, which use representative input data to guide quantization decisions, and calibration-free approaches, which determine quantization parameters entirely from weight statistics. Calibration-based methods such as AWQ~\cite{awq} and GPTQ~\cite{gptq} are most commonly studied in autoregressive LLM settings, where representative calibration data is naturally defined in terms of text prompts. Prior ASR work has also explored low-bit quantization-aware training, including 2-bit QAT for Conformer models on LibriSpeech and large-scale internal data~\cite{2bitqat}, though such approaches require retraining. For a stateful streaming transducer operating on audio input, defining representative calibration data and chunk-level execution conditions is less straightforward, so in this work we focus on calibration-free post-training strategies that require no additional training or calibration data.

All quantization variants use \emph{weight-only} block quantization: each weight matrix $W$ is partitioned into contiguous blocks of $b$ elements (block size $b=32$ in our experiments), and each block is independently mapped to $n$-bit integers ($n \in \{4, 8\}$) with a per-block scale $s$ and zero-point $z$. Activations remain in FP32 at inference time; the quantized weights are consumed by the fused ONNX \texttt{MatMulNBits} operator, which combines linear dequantization with matrix multiplication. We evaluate two calibration-free quantization algorithms that differ in how $s$ and $z$ are computed:

\paragraph{Round-To-Nearest (RTN).} For each block with weights $\{w_j\}_{j=1}^{b}$, RTN computes scale and zero-point directly from the weight range:
\[
s = \frac{w_{\max} - w_{\min}}{2^n - 1}, \qquad z = \mathrm{round}\!\left(\frac{-w_{\min}}{s}\right)
\]
Each weight is then quantized in a single pass:
\[
q_j = \mathrm{clamp}\!\left(\mathrm{round}\!\left(\frac{w_j}{s} + z\right),\; 0,\; 2^n - 1\right)
\]
RTN is one of the fastest quantization methods, but provides no mechanism to minimize the resulting quantization error.

\paragraph{K-quant.} K-quant improves upon RTN by optimizing the scale and offset per block to minimize reconstruction error, with greater emphasis on preserving large-magnitude weights. To our knowledge, this quantization scheme has limited formal description in prior literature, so we summarize the procedure used in our implementation.

Each weight $w_j$ in a block is assigned an importance weight $\alpha_j$ that combines the block's root-mean-square with the element's own magnitude:
\[
\alpha_j = \underbrace{\sqrt{\frac{1}{b}\sum_{k=1}^{b} w_k^2}}_{\text{block RMS (constant)}} + \;|w_j|
\]

Given fixed quantized integers $q_j$ (obtained from an initial RTN-like pass), k-quant then solves for the optimal affine mapping $(s^*, m^*)$ such that the dequantized weights $\hat{w}_j = s^* \cdot q_j + m^*$ are as close as possible to the original weights $w_j$:
\[
\min_{s^*,\, m^*} \sum_{j=1}^{b} \alpha_j \left(\underbrace{s^* \cdot q_j + m^*}_{\hat{w}_j \text{ (reconstructed)}} -\, w_j\right)^2
\]
This is a two-variable weighted least-squares problem with a closed-form solution. The block RMS provides a scale-adaptive baseline importance for all elements, while $|w_j|$ ensures that larger weights receive proportionally more attention during optimization. The RMS term is necessary because using $|w_j|$ alone would assign near-zero importance to near-zero weights, allowing the optimizer to introduce arbitrary error on them. Since even a small weight encodes meaningful information (e.g., suppressing a particular output dimension), this would degrade model quality. The RMS term sets a minimum importance floor that adapts to the block's overall magnitude scale, preventing any weight from being completely ignored.

To explore beyond the initial quantization, our k-quant evaluates 20 candidate scale factors, uniformly spaced in a narrow range around the initial RTN scale, each producing different integer assignments $q_j$. For each candidate, the optimal $(s^*, m^*)$ is recomputed in closed form, and the candidate with the lowest weighted error is kept.

\medskip
Quantization is applied only to the encoder, which accounts for over 95\% of model parameters. The decoder (LSTM prediction network) and joiner remain in FP32: their combined size is under 35\,MB, so quantizing them would yield negligible size savings while risking degradation in the RNNT decoding loop, where the joiner is invoked at every encoder time step. Similarly, the streaming cache tensors (channel and temporal context) are maintained in FP32 to preserve numerical stability across chunk boundaries.

\medskip
We evaluated the following quantization configurations:

\begin{itemize}
    \item \textbf{int4 RTN:} 4-bit RTN quantization. The simplest and fastest method, serving as a lower bound on quantization quality.
    \item \textbf{int8 k-quant:} 8-bit k-quant quantization applied to all quantizable layers.
    \item \textbf{int4 k-quant:} Uniform 4-bit k-quant applied to all quantizable layers.
    \item \textbf{int4-mixed k-quant:} Mixed-precision k-quant where most linear layers use int4, but layers identified as accuracy-sensitive (e.g., attention Q/K/V/O projections in all encoder layers, and the first and last encoder layers) retain int8 precision.
    \item \textbf{int4 k-quant + ConvInteger/MatMulInteger:} int4 k-quant with additional graph-level optimization that replaces floating-point Conv and MatMul operators with their integer-arithmetic counterparts (ConvInteger, MatMulInteger). While this can improve throughput on certain hardware, it performs the entire computation in integer arithmetic rather than dequantizing back to floating point, accumulating rounding errors through the network.
\end{itemize}

% ============================================================
\section{Results}
\label{sec:results}

\subsection{Quantization Results}

Table~\ref{tab:quantization} reports the full evaluation results across all quantization variants.

\begin{table}[H]
\centering
\caption{Nemotron-0.6B quantization results. Streaming config: (7,10,7), 0.56\,s delay, 5.6\,s history. CPU inference, batch\_size=1.}
\label{tab:quantization}
\resizebox{\textwidth}{!}{%
\begin{tabular}{@{}llcccccccccccc@{}}
\toprule
\textbf{Variant} & \textbf{Format} & \textbf{Size} & \textbf{Device} & \textbf{Avg} & \textbf{RTFx} & \textbf{AMI} & \textbf{Earn.} & \textbf{Giga} & \textbf{LS-c} & \textbf{LS-o} & \textbf{SPGI} & \textbf{TED} & \textbf{VoxP.} \\
\midrule
\textit{Baseline (PyTorch)} & \textit{PyTorch} & \textit{2.47\,GB} & \textit{CUDA} & \textit{7.28} & \textit{--} & \textit{11.80} & \textit{12.64} & \textit{11.50} & \textit{2.33} & \textit{5.08} & \textit{2.72} & \textit{4.46} & \textit{7.69} \\
FP32 & ONNX & 2.47\,GB & CPU & 8.03 & 6.73 & 16.40 & 13.32 & 12.00 & 2.35 & 5.01 & 2.61 & 4.66 & 7.90 \\
int8 k-quant & ONNX & 1.28\,GB & CPU & 8.01 & 7.25 & 16.37 & 13.35 & 11.97 & 2.36 & 4.97 & 2.52 & 4.62 & 7.92 \\
int4-mixed k-quant & ONNX & 0.73\,GB & CPU & 8.12 & 7.15 & 16.72 & 13.52 & 12.08 & 2.36 & 5.02 & 2.64 & 4.70 & 7.88 \\
int4 k-quant & ONNX & 0.67\,GB & CPU & 8.20 & 7.20 & 17.05 & 13.60 & 12.10 & 2.38 & 5.04 & 2.83 & 4.65 & 7.98 \\
int4 k-quant w/ ConvInt & ONNX & 0.67\,GB & CPU & 10.14 & 8.00 & 20.55 & 11.68 & 14.73 & 5.32 & 10.14 & 3.05 & 5.64 & 9.98 \\
int4 RTN & ONNX & 0.66\,GB & CPU & 8.46 & 7.30 & 18.94 & 13.52 & 12.12 & 2.42 & 5.10 & 2.85 & 4.72 & 8.05 \\
\bottomrule
\end{tabular}%
}
\end{table}

\begin{figure}[H]
\centering
\includegraphics[width=0.85\textwidth]{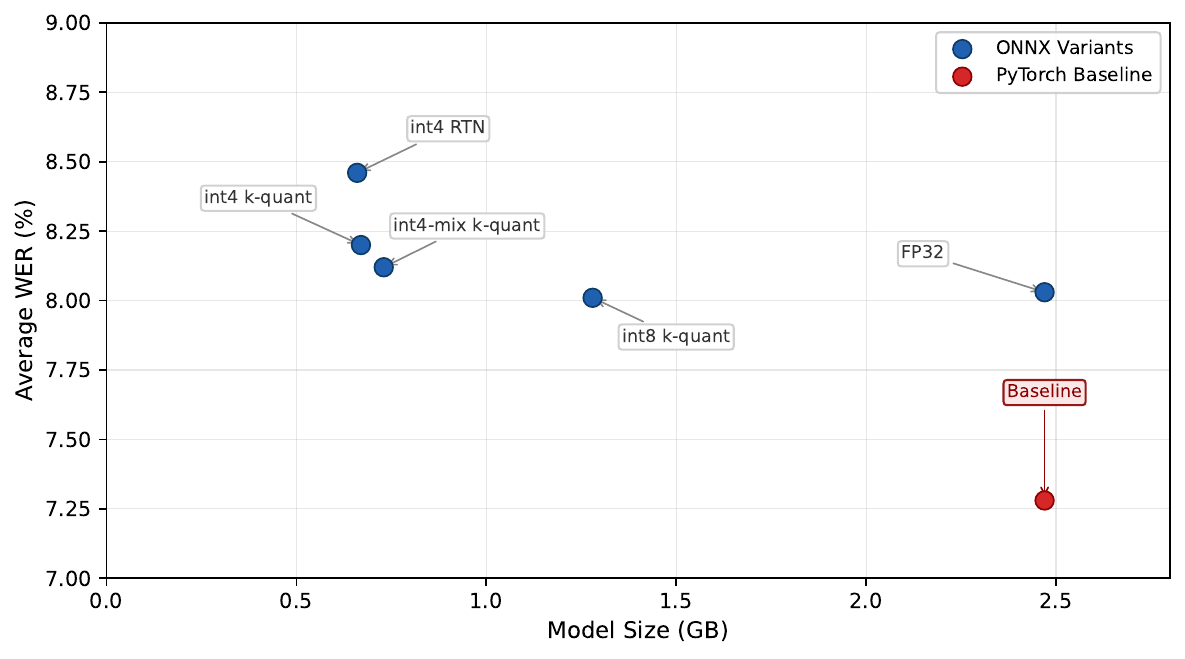}
\caption{Model size vs.\ WER for Nemotron quantization variants. The int4 k-quant variant achieves 8.20\% WER at 0.67\,GB, within 0.17\% of the ONNX FP32 baseline (8.03\%).}
\label{fig:size_wer}
\end{figure}

\begin{figure}[H]
\centering
\includegraphics[width=0.85\textwidth]{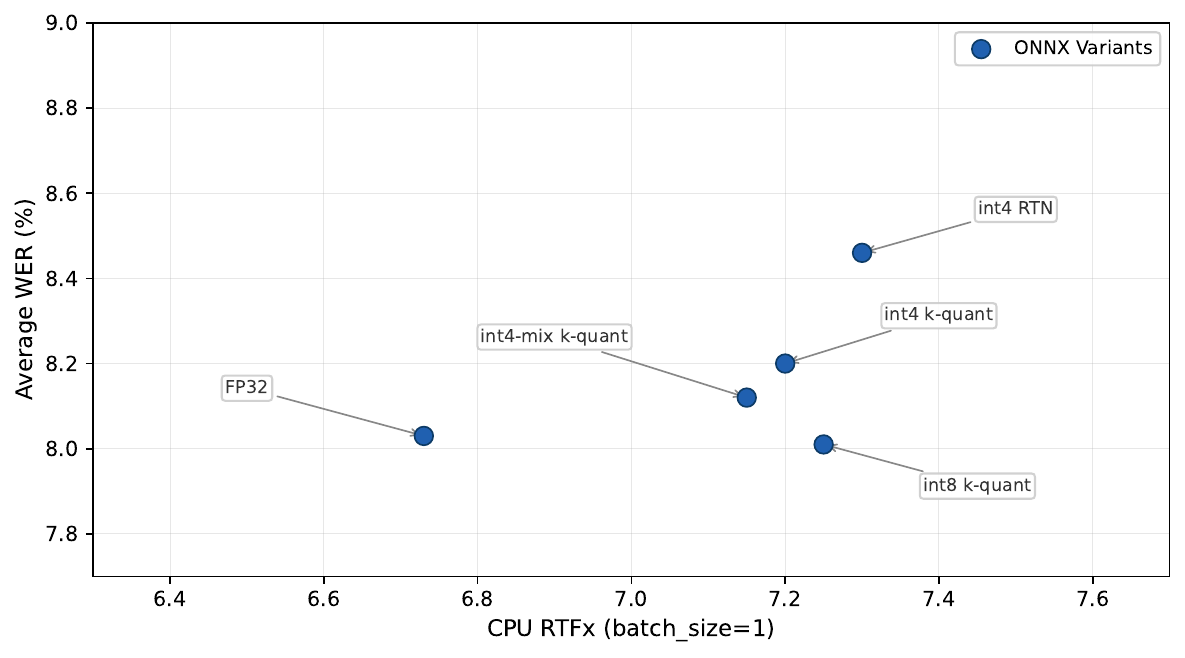}
\caption{CPU RTFx vs.\ WER. All ONNX variants achieve RTFx $>6\times$ real-time on CPU, with quantized variants slightly faster than FP32.}
\label{fig:rtfx_wer}
\end{figure}

\subsection{Key Findings}

\begin{enumerate}
    \item \textbf{Quantization preserves accuracy remarkably well.} The int8 k-quant variant achieves 8.01\% average WER, essentially matching the ONNX FP32 baseline (8.03\%), with a 48\% reduction in model size. Even at int4 precision with a model size of just 0.67\,GB (73\% smaller), the WER degrades by only 0.17\% absolute (2.1\% relative), demonstrating that aggressive 4-bit compression is viable for streaming ASR.
    
    \item \textbf{CPU inference is practical.} All ONNX variants achieve RTFx $> 6\times$ on CPU, meaning they can process audio more than 6 times faster than real-time---well within the requirements for streaming applications. With the selected (7, 10, 7) streaming configuration (0.56\,s algorithmic delay), this yields an effective time-to-first-token well under 0.7\,s, dominated by audio accumulation rather than inference compute.
    
    \item \textbf{Quantization accelerates inference.} The quantized variants (int8, int4-mixed, int4) all achieve slightly higher RTFx than FP32 ONNX (7.15--7.30$\times$ vs.\ 6.7$\times$), showing that reduced precision translates to faster throughput on CPU.
    
    \item \textbf{ONNX FP32 vs.\ PyTorch baseline gap.} The ONNX FP32 variant shows a 0.75\% absolute WER increase over the PyTorch CUDA baseline (8.03\% vs.\ 7.28\%). We suspect that this discrepancy is not caused by the ONNX export itself, since the graph is numerically equivalent, but rather by differences in kernel implementations that may compound through 24 conformer layers.
\end{enumerate}

% ============================================================
\subsection{Trade-off of Algorithmic Delay and Effective Latency}
\label{sec:latency_tradeoff}

\begin{figure}[H]
\centering
\includegraphics[width=0.85\textwidth]{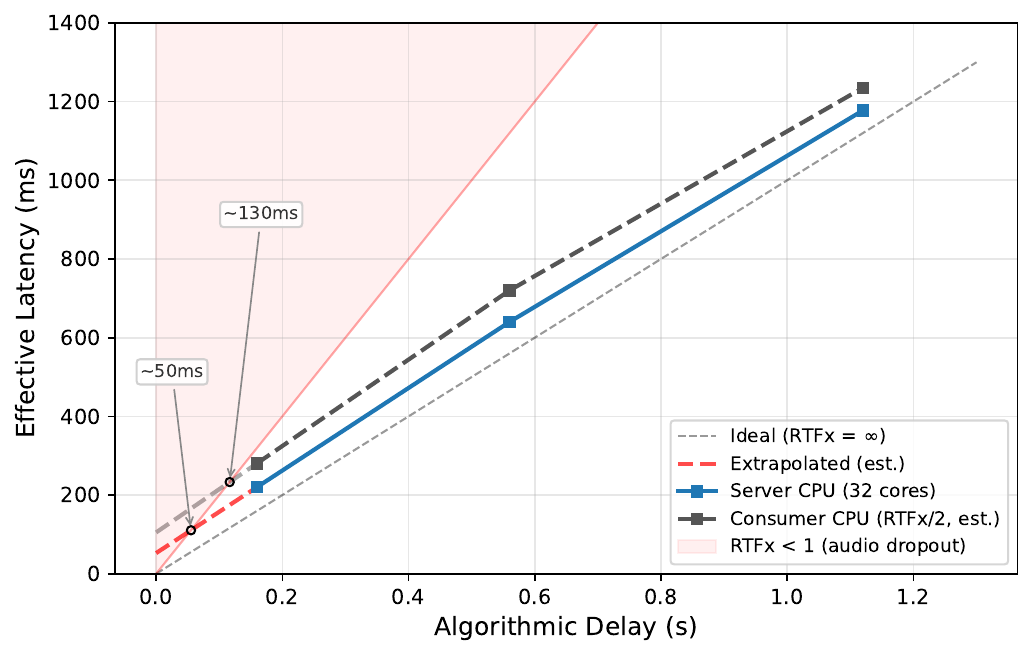}
\caption{Effective latency vs.\ algorithmic delay for Nemotron-0.6B int4 k-quant on CPU. The red shaded region indicates where RTFx drops below 1$\times$ (audio dropout).}
\label{fig:latency_rtfx}
\end{figure}

RTFx varies significantly with the streaming configuration. Smaller chunk sizes reduce algorithmic delay but also reduce per-chunk compute efficiency, as fixed per-invocation overhead (session invoke, tensor binding, cache management) is amortized over less audio. Figure~\ref{fig:latency_rtfx} illustrates this trade-off using measured RTFx values from the int4 k-quant model across three streaming configurations (0.16\,s, 0.56\,s, and 1.12\,s algorithmic delay): the effective latency (algorithmic delay plus compute time) diverges from the ideal as chunk size decreases. Extrapolating from the measured trend suggests that configurations near $\sim$50\,ms algorithmic delay may still be feasible on our server CPU (32 cores), though we did not evaluate such settings directly. On weaker consumer hardware, the margin is expected to shrink. For example, under an illustrative assumption of roughly 2$\times$ lower throughput than our server, the 0.16\,s configuration would operate at approximately RTFx $\approx$ 1.3$\times$, leaving limited headroom for background load, while substantially smaller chunk sizes could risk audio dropout.

\medskip
The 0.16\,s configuration achieves 8.89\% average WER with the ONNX int4 k-quant model on CPU (see Table~\ref{tab:app_quant_160ms} in the appendix), only 0.69\% above the 0.56\,s result, while reducing algorithmic delay by more than $3\times$. These results suggest that further latency reduction beyond our recommended 0.56\,s configuration is feasible for applications that prioritize responsiveness, provided the target hardware offers sufficient compute headroom.

% ============================================================
\section{Recommendation and Selected Configuration}
\label{sec:recommendation}

Based on our evaluation, we selected the Nemotron-0.6B int4 k-quant variant with the (7, 10, 7) streaming configuration and 0.56\,s algorithmic delay as our recommended on-device streaming ASR model. The optimized model is available through Microsoft's Foundry Local platform, with streaming inference support across C\#, Python, JavaScript, C++, and Rust via the ONNX Runtime GenAI SDK.

% ============================================================
\section{Discussion}
\label{sec:discussion}

Our evaluation of over 50 configurations across six model families reveals several noteworthy patterns worth discussing. First, there is a clear architectural divide between models designed for batch processing and those designed for streaming. Batch-oriented models such as Qwen3-ASR-1.7B and Whisper achieve strong offline WERs but degrade substantially when adapted to chunked or streaming operation. Qwen3-ASR-1.7B, for example, jumps from 5.90\% to 10.45\% WER when chunked at 2.4\,s stride. In contrast, Nemotron-0.6B's cache-aware architecture, purpose-built for streaming, loses only 0.21\% absolute when moving from batch to its optimal streaming configuration. More broadly, our results suggest that offline ASR quality is a poor proxy for low-latency streaming quality: despite strong batch WER, several modern models exhibit substantial degradation when constrained to practical streaming settings.

The sensitivity to history context is also striking. Comparing the (7, 10, 7) and (7, 2, 7) configurations---identical except for left context---shows a 1.23\% WER gap (7.28\% vs.\ 8.51\%), underscoring that streaming models need sufficient lookback to maintain accuracy. This has direct implications for memory budgeting on edge devices: the 5.6\,s history window required by the optimal configuration is a modest but non-negligible cost.

Quantization preserves accuracy remarkably well. On the full evaluation sets, the int4 k-quant variant achieves 8.20\% average WER compared to the 8.03\% ONNX FP32 baseline, a degradation of only 0.17\% absolute despite a 73\% reduction in model size (from 2.47\,GB to 0.67\,GB). The degradation is distributed across most datasets, with the largest increases on AMI (+0.65\%) and SPGISpeech (+0.22\%), while TED-LIUM actually shows no degradation (4.65\% vs.\ 4.66\%). The int8 variant at 1.28\,GB essentially matches FP32 accuracy (8.01\% vs.\ 8.03\%), confirming that 8-bit quantization is effectively lossless for this architecture. One notable negative result is the ConvInteger/MatMulInteger variant, which degrades to 10.14\% WER despite using the same int4 k-quant weights. Unlike standard quantized operators that dequantize inputs back to floating point before computing, ConvInteger and MatMulInteger perform the entire computation in integer arithmetic, accumulating rounding errors through the encoder's 24 conformer layers of interleaved convolutions and attention.

From a resource utilization perspective, the ONNX Nemotron variants occupy a compelling position: all variants achieve RTFx $>6\times$ on CPU, running over $6\times$ faster than real-time. Other evaluated architectures exhibited higher WER or latency under the tested streaming configurations.

\subsection{Practical Scope and Positioning}

The results in this report show that compact on-device ASR can achieve strong core transcription quality on standard English benchmarks while operating with low latency and no GPU requirement. More broadly, the observed ESB performance suggests that compact on-device streaming ASR is narrowing the historical quality gap to production systems on core transcription benchmarks. However, these results should not be over-interpreted. While the ESB suite covers several challenging domains, including far-field meetings (AMI) and telephony-quality financial calls (Earnings22), it represents only a limited subset of real-world speech recognition use cases. Our evaluation focuses on core transcription accuracy and does not cover several important system-level capabilities often required in production settings, such as robust inverse text normalization (numbers, dates, currencies), speaker diarization, code-switching, custom vocabulary adaptation, or broader internal evaluation suites designed to reflect customer usage. The ESB results should therefore be understood as measuring core recognition quality across a representative but limited set of domains, rather than as a comprehensive assessment of end-to-end speech recognition system quality. In this context, compact on-device ASR is best viewed as a strong option for privacy-preserving, offline, and latency-sensitive scenarios.

\subsection{Concurrent Work}

We note that concurrent to this work, Mistral released Voxtral Mini~\cite{voxtral}, a multimodal model with strong ASR capabilities including streaming support. While promising, its model size ($\sim$4\,GB) exceeds the sub-1\,GB target for our edge deployment scenario. Similarly, Useful Sensors released Moonshine v2~\cite{moonshine}, a lightweight encoder-decoder model with sliding-window attention designed for on-device streaming. Although its model variants fit comfortably within our size constraints, its autoregressive design introduces a different latency profile from the transducer-based system studied here, with output latency that can grow with transcript length even when time-to-first-token is low. In addition, as with other sequence-to-sequence ASR models, it may be more susceptible to repetitions or hallucination-like behaviors in challenging conditions. A comprehensive empirical comparison with these models is left for future work.
\subsection{Future Work}

The current release focuses on English-only streaming ASR. We plan to extend this work along two key axes:

\begin{itemize}
    \item \textbf{Multilingual support.} Expanding beyond English to additional languages, leveraging multilingual base models and language-specific quantization calibration to maintain quality across diverse languages while preserving the compact edge deployment profile.
    \item \textbf{Broader model support.} Integrating additional ASR architectures, enabling users to select models based on their specific accuracy, latency, and resource requirements.
\end{itemize}

\newpage
\appendix

\section{Complete Evaluation Results}
\label{app:results}

This appendix reports all configurations evaluated during this study, including incomplete runs and configurations not featured in the main text. All WER values are in percent (\%).

\subsection{Whisper --- All Configurations}

\begin{table}[H]
\centering
\caption{All Whisper configurations evaluated.}
\label{tab:app_whisper}
\resizebox{\textwidth}{!}{%
\begin{tabular}{@{}llccccccccccc@{}}
\toprule
\textbf{Model} & \textbf{Format} & \textbf{Mode} & \textbf{Size} & \textbf{Avg} & \textbf{AMI} & \textbf{Earn.} & \textbf{Giga} & \textbf{LS-c} & \textbf{LS-o} & \textbf{SPGI} & \textbf{TED} & \textbf{VoxP.} \\
\midrule
whisper-large-v3-turbo & PyTorch & Batch & 1.62\,GB & 7.83 & 16.13 & 11.63 & 10.14 & 2.10 & 4.24 & 2.97 & 3.57 & 11.87 \\
whisper-large-v3-turbo & ORT FP16 & Batch & 1.75\,GB & 7.52 & 16.36 & 11.38 & 10.13 & 2.17 & 4.24 & 2.93 & 3.62 & 9.30 \\
whisper-small.en & PyTorch & Batch & 0.97\,GB & (8.59) & 17.93 & 12.97 & 11.35 & 3.05 & 7.25 & 3.60 & 4.07 & 8.50 \\
faster-whisper (small) & PyTorch & Chunk (3\,s) & 0.97\,GB & 24.74 & 32.90 & 32.59 & -- & 17.23 & 22.72 & 19.81 & -- & 23.18 \\
\bottomrule
\end{tabular}%
}
\end{table}

\subsection{Nemotron-0.6B --- All Streaming Configurations}

\begin{table}[H]
\centering
\caption{All Nemotron-0.6B streaming configurations (PyTorch, CUDA, batch\_size=16).}
\label{tab:app_nemotron}
\resizebox{\textwidth}{!}{%
\begin{tabular}{@{}lcccccccccccc@{}}
\toprule
\textbf{Config} & \textbf{Delay} & \textbf{History} & \textbf{Avg} & \textbf{RTFx} & \textbf{AMI} & \textbf{Earn.} & \textbf{Giga} & \textbf{LS-c} & \textbf{LS-o} & \textbf{SPGI} & \textbf{TED} & \textbf{VoxP.} \\
\midrule
Batch (offline) & -- & -- & 7.07 & 990.48 & 11.16 & 12.38 & 11.35 & 2.28 & 4.83 & 2.63 & 4.42 & 7.48 \\
(70, 2, 70) & 5.6\,s & 11.2\,s & 7.10 & 268.39 & 11.31 & 12.46 & 11.39 & 2.28 & 4.84 & 2.64 & 4.42 & 7.47 \\
(14, 2, 14) & 1.12\,s & 2.24\,s & 7.46 & 260.05 & 11.38 & 12.49 & 11.39 & 2.41 & 5.07 & 2.71 & 4.50 & 9.75 \\
(7, 10, 7) & 0.56\,s & 5.6\,s & 7.28 & 145.16 & 11.80 & 12.64 & 11.50 & 2.33 & 5.08 & 2.72 & 4.46 & 7.69 \\
(7, 2, 7) & 0.56\,s & 1.12\,s & 8.51 & 144.71 & 11.93 & 13.93 & 11.76 & 3.43 & 6.00 & 3.47 & 5.11 & 12.49 \\
(2, 35, 2) & 0.16\,s & 5.6\,s & 8.04 & 45.05 & 14.82 & 13.66 & 12.44 & 2.49 & 5.52 & 2.90 & 4.72 & 7.79 \\
(2, 2, 2) & 0.16\,s & 0.32\,s & -- & -- & 19.75 & 26.36 & 19.46 & 11.86 & 17.88 & -- & 13.32 & 32.42 \\
(1, 70, 1) & 0.08\,s & 5.6\,s & 8.91 & -- & 17.76 & 14.64 & 13.33 & 2.61 & 5.93 & 3.73 & 5.06 & 8.24 \\
\bottomrule
\end{tabular}%
}
\end{table}

\subsection{Parakeet TDT-0.6B-v3 --- All Chunking Configurations}

\begin{table}[H]
\centering
\caption{All Parakeet TDT-0.6B-v3 configurations. Batch and chunked modes (CUDA, batch\_size=16). Config format: delay (left-current-right = total context).}
\label{tab:app_parakeet}
\resizebox{\textwidth}{!}{%
\begin{tabular}{@{}llccccccccccc@{}}
\toprule
\textbf{Config} & \textbf{Precision} & \textbf{Avg} & \textbf{RTFx} & \textbf{AMI} & \textbf{Earn.} & \textbf{Giga} & \textbf{LS-c} & \textbf{LS-o} & \textbf{SPGI} & \textbf{TED} & \textbf{VoxP.} \\
\midrule
Batch & -- & (6.32) & (3332.74) & (11.39) & (11.19) & (9.57) & (1.92) & (3.59) & (3.98) & (2.80) & (6.09) \\
\midrule
28.8\,s (7.2-21.6-7.2=36) & bf16 & 16.94 & 211.45 & 41.64 & 32.38 & 20.56 & 3.69 & 6.10 & 8.63 & 10.73 & 11.75 \\
13.2\,s (8.4-11.6-1.6=21.6) & bf16 & 9.24 & 272.39 & 15.96 & 17.37 & 12.10 & 2.52 & 4.34 & 7.40 & 5.65 & 8.59 \\
11.6\,s (8.4-10.4-1.2=20) & bf16 & 12.13 & 358.98 & 22.25 & 22.61 & 18.23 & 3.76 & 4.82 & 8.41 & 8.36 & 8.62 \\
10.8\,s (8.4-10-0.8=19.2) & bf16 & 16.10 & 307.51 & 37.21 & 30.55 & 21.74 & 4.17 & 5.32 & 9.14 & 10.40 & 10.32 \\
12.8\,s (3.2-9.6-3.2=16) & bf16 & 9.66 & 257.87 & 15.56 & 17.80 & 12.39 & 3.69 & 5.17 & 7.86 & 6.16 & 8.66 \\
10.4\,s (5.6-8.8-1.6=16) & bf16 & 9.22 & 250.41 & 15.35 & 17.12 & 12.02 & 3.44 & 5.27 & 7.23 & 5.32 & 8.01 \\
9.6\,s (5.6-8-1.6=15.2) & bf16 & 9.33 & 361.64 & 15.26 & 17.36 & 12.15 & 3.59 & 5.12 & 7.41 & 5.65 & 8.10 \\
9.6\,s (5.6-8-1.6=15.2) & fp16 & 9.59 & 219.76 & 15.27 & 19.44 & 12.16 & 3.47 & 5.26 & 7.41 & 5.61 & 8.07 \\
9.2\,s (5.6-7.6-1.6=14.8) & bf16 & 11.61 & 402.68 & 21.72 & 21.74 & 17.68 & 3.35 & 4.88 & 7.97 & 7.36 & 8.14 \\
8.8\,s (5.6-8-0.8=14.4) & bf16 & 15.83 & 257.07 & 35.91 & 29.09 & 20.56 & 6.06 & 6.84 & 8.63 & 9.56 & 9.95 \\
8.8\,s (5.6-7.6-1.2=14.4) & bf16 & 15.99 & 204.10 & 36.08 & 29.15 & 21.49 & 5.07 & 6.94 & 8.66 & 10.56 & 9.98 \\
8\,s (5.6-7.2-0.8=13.6) & bf16 & 15.97 & 404.98 & 35.48 & 29.27 & 21.84 & 5.69 & 6.39 & 8.89 & 10.15 & 10.04 \\
9.6\,s (2.4-7.2-2.4=12) & fp16 & 12.55 & 323.81 & 21.65 & 21.53 & 18.32 & 6.43 & 7.37 & 8.46 & 7.95 & 8.71 \\
9.6\,s (2.4-7.2-2.4=12) & bf16 & 16.46 & 361.34 & 35.60 & 29.04 & 21.92 & 7.45 & 8.09 & 8.92 & 10.45 & 10.22 \\
2.4\,s (9.6-0.8-1.6=12) & bf16 & 12.83 & 83.68 & 16.73 & 19.81 & 15.27 & 6.63 & 8.25 & 17.82 & 9.95 & 8.19 \\
6.4\,s (1.6-4.8-1.6=8) & fp16 & 11.54 & 228.68 & 15.51 & 18.91 & 13.45 & 9.27 & 9.73 & 9.39 & 7.20 & 8.82 \\
6.4\,s (1.6-4.8-1.6=8) & bf16 & 11.52 & -- & 15.52 & 18.91 & 13.44 & 9.34 & 9.57 & 9.33 & 7.26 & 8.79 \\
3.2\,s (0.8-2.4-0.8=4) & bf16 & 32.17 & 209.69 & 43.08 & 38.13 & 35.66 & 35.62 & 30.13 & 31.28 & 30.93 & 12.55 \\
\bottomrule
\end{tabular}%
}
\end{table}

\subsection{Parakeet TDT-0.6B-v3 --- ONNX Quantization}

\begin{table}[H]
\centering
\caption{Parakeet TDT-0.6B-v3 ONNX quantization variants (batch mode, CUDA, batch\_size=16).}
\label{tab:app_parakeet_quant}
\resizebox{\textwidth}{!}{%
\begin{tabular}{@{}lccccccccccc@{}}
\toprule
\textbf{Format} & \textbf{Size} & \textbf{Avg} & \textbf{RTFx} & \textbf{AMI} & \textbf{Earn.} & \textbf{Giga} & \textbf{LS-c} & \textbf{LS-o} & \textbf{SPGI} & \textbf{TED} & \textbf{VoxP.} \\
\midrule
ONNX fp16 & 1.30\,GB & -- & -- & 21.46 & -- & -- & 2.21 & 3.96 & -- & -- & 6.82 \\
ONNX int8 & 1.02\,GB & 9.92 & 97.10 & 18.82 & 21.19 & 12.92 & 2.18 & 4.25 & 7.17 & 5.79 & 7.07 \\
ONNX int4 & 0.74\,GB & 11.50 & 112.94 & 21.22 & 26.34 & 14.39 & 2.16 & 4.02 & 8.97 & 8.17 & 6.77 \\
\bottomrule
\end{tabular}%
}
\end{table}

\subsection{Canary-1B-v2 --- All Configurations}

\begin{table}[H]
\centering
\caption{Canary-1B-v2 configurations (PyTorch bf16, CUDA, batch\_size=16).}
\label{tab:app_canary}
\resizebox{\textwidth}{!}{%
\begin{tabular}{@{}llccccccccccc@{}}
\toprule
\textbf{Model} & \textbf{Config} & \textbf{Size} & \textbf{Avg} & \textbf{RTFx} & \textbf{AMI} & \textbf{Earn.} & \textbf{Giga} & \textbf{LS-c} & \textbf{LS-o} & \textbf{SPGI} & \textbf{TED} & \textbf{VoxP.} \\
\midrule
Canary-1B-v2 & Batch & 6.36\,GB & (7.15) & (749) & (16.01) & (11.79) & (10.82) & (2.18) & (3.56) & (2.28) & (4.29) & (6.25) \\
Canary-1B-v2 & Chunk 4.8\,s & 6.36\,GB & 12.45 & -- & 22.10 & 22.04 & 14.60 & 4.87 & 5.96 & 4.42 & 7.64 & 17.98 \\
\bottomrule
\end{tabular}%
}
\par\noindent{\scriptsize Chunked config: (10.0-2.4-2.4=14.8) with AlignAtt streaming policy.}
\end{table}

\subsection{Conformer Transducer XL --- All Configurations}

\begin{table}[H]
\centering
\caption{Conformer Transducer XL chunked configurations (PyTorch bf16, CUDA, batch\_size=16). Config: delay (left-current-right = total context).}
\label{tab:app_conformer}
\resizebox{\textwidth}{!}{%
\begin{tabular}{@{}lccccccccccc@{}}
\toprule
\textbf{Config} & \textbf{Avg} & \textbf{RTFx} & \textbf{AMI} & \textbf{Earn.} & \textbf{Giga} & \textbf{LS-c} & \textbf{LS-o} & \textbf{SPGI} & \textbf{TED} & \textbf{VoxP.} \\
\midrule
6.4\,s (1.6-4.8-1.6=8) & 10.78 & 222.90 & 22.07 & 23.47 & 14.21 & 2.22 & 4.10 & 7.93 & 5.48 & 6.80 \\
3.2\,s (4.8-1.6-1.6=8) & 10.96 & 133.69 & 22.31 & 24.22 & 13.99 & 2.04 & 3.78 & 7.86 & 5.40 & 8.05 \\
2.4\,s (5.6-0.8-1.6=8) & 11.06 & 82.03 & 22.57 & 24.44 & 14.05 & 2.11 & 3.83 & 7.95 & 5.36 & 8.18 \\
2.4\,s (9.6-0.8-1.6=12) & 11.02 & 44.28 & 22.69 & 24.47 & 13.97 & 1.93 & 3.46 & 7.86 & 5.26 & 8.49 \\
\bottomrule
\end{tabular}%
}
\end{table}

\subsection{Qwen3-ASR --- All Configurations}

\begin{table}[H]
\centering
\caption{Qwen3-ASR configurations (PyTorch, CUDA, batch\_size=32).}
\label{tab:app_qwen}
\resizebox{\textwidth}{!}{%
\begin{tabular}{@{}llccccccccccc@{}}
\toprule
\textbf{Model} & \textbf{Config} & \textbf{Avg} & \textbf{RTFx} & \textbf{AMI} & \textbf{Earn.} & \textbf{Giga} & \textbf{LS-c} & \textbf{LS-o} & \textbf{SPGI} & \textbf{TED} & \textbf{VoxP.} \\
\midrule
Qwen3-ASR-1.7B & Batch & 5.90 & 187.91 & 11.76 & 10.26 & 8.75 & 1.60 & 3.41 & 2.83 & 2.28 & 6.34 \\
Qwen3-ASR-0.6B & Batch & 6.69 & 194.71 & 13.77 & 11.03 & 9.16 & 2.12 & 4.47 & 3.04 & 2.83 & 7.09 \\
\midrule
Qwen3-ASR-1.7B & Chunk 3\,s (1\,s stride) & 9.69 & 61.70 & 15.54 & 14.94 & 11.28 & 3.58 & 5.99 & 5.26 & 5.03 & 9.90 \\
Qwen3-ASR-0.6B & Chunk 3\,s (1\,s stride) & 9.81 & 64.11 & 17.47 & 15.05 & 11.37 & 3.94 & 7.02 & 5.22 & 5.02 & 13.41 \\
Qwen3-ASR-1.7B & Chunk 2.4\,s (1.2\,s stride) & 10.45 & 42.76 & 16.97 & 16.92 & 12.29 & 4.95 & 7.69 & 6.70 & 6.34 & 11.75 \\
Qwen3-ASR-0.6B & Chunk 2.4\,s (1.2\,s stride) & 10.41 & 46.32 & 18.55 & 16.60 & 12.00 & 4.97 & 8.31 & 6.13 & 5.75 & 10.99 \\
\bottomrule
\end{tabular}%
}
\end{table}

\subsection{Cross-Model Streaming Comparison (CPU)}

\begin{table}[H]
\centering
\caption{Cross-model streaming comparison (CPU, batch\_size=1). Best streaming or low-latency configuration per model.}
\label{tab:app_streaming_cpu}
\resizebox{\textwidth}{!}{%
\begin{tabular}{@{}lccccccccccccc@{}}
\toprule
\textbf{Model} & \textbf{Size} & \textbf{Delay} & \textbf{Avg} & \textbf{BSF} & \textbf{RTFx} & \textbf{AMI} & \textbf{Earn.} & \textbf{Giga} & \textbf{LS-c} & \textbf{LS-o} & \textbf{SPGI} & \textbf{TED} & \textbf{VoxP.} \\
\midrule
Nemotron-0.6B & 2.47\,GB & 0.56\,s & 7.28 & \textbf{1.03} & 2.46 & 11.80 & 12.64 & 11.50 & 2.33 & 5.08 & 2.72 & 4.46 & 7.69 \\
Qwen3-ASR-1.7B & 4.70\,GB & 2.4\,s & 10.45 & 1.77 & 0.49 & 16.97 & 16.92 & 12.29 & 4.95 & 7.69 & 6.70 & 6.34 & 11.75 \\
Parakeet TDT-0.6B-v3 & 2.51\,GB & 2.4\,s & 12.83 & 2.03 & 1.38 & 16.73 & 19.81 & 15.27 & 6.63 & 8.25 & 17.82 & 9.95 & 8.19 \\
Conformer Trans. XL & 2.58\,GB & 2.4\,s & 11.06 & -- & 1.27 & 22.57 & 24.44 & 14.05 & 2.11 & 3.83 & 7.95 & 5.36 & 8.18 \\
Canary-1B-v2 & 6.36\,GB & 4.8\,s & 12.45 & 1.74 & 2.93 & 22.10 & 22.04 & 14.60 & 4.87 & 5.96 & 4.42 & 7.64 & 17.98 \\
\bottomrule
\end{tabular}%
}
\end{table}

\subsection{Nemotron-0.6B --- ONNX Quantization Results}

\begin{table}[H]
\centering
\caption{Nemotron-0.6B ONNX quantization results. Config: (7,10,7), 0.56\,s delay, 5.6\,s history. CPU inference, batch\_size=1.}
\label{tab:app_quant}
\resizebox{\textwidth}{!}{%
\begin{tabular}{@{}llcccccccccccc@{}}
\toprule
\textbf{Variant} & \textbf{Format} & \textbf{Size} & \textbf{Device} & \textbf{Avg} & \textbf{RTFx} & \textbf{AMI} & \textbf{Earn.} & \textbf{Giga} & \textbf{LS-c} & \textbf{LS-o} & \textbf{SPGI} & \textbf{TED} & \textbf{VoxP.} \\
\midrule
\textit{Baseline (PyTorch)} & \textit{PyTorch} & \textit{2.47\,GB} & \textit{CUDA} & \textit{7.28} & \textit{--} & \textit{11.80} & \textit{12.64} & \textit{11.50} & \textit{2.33} & \textit{5.08} & \textit{2.72} & \textit{4.46} & \textit{7.69} \\
FP32 & ONNX & 2.47\,GB & CPU & 8.03 & 6.73 & 16.40 & 13.32 & 12.00 & 2.35 & 5.01 & 2.61 & 4.66 & 7.90 \\
int8 k-quant & ONNX & 1.28\,GB & CPU & 8.01 & 7.25 & 16.37 & 13.35 & 11.97 & 2.36 & 4.97 & 2.52 & 4.62 & 7.92 \\
int4-mixed k-quant & ONNX & 0.73\,GB & CPU & 8.12 & 7.15 & 16.72 & 13.52 & 12.08 & 2.36 & 5.02 & 2.64 & 4.70 & 7.88 \\
\textbf{int4 k-quant} & ONNX & \textbf{0.67\,GB} & CPU & \textbf{8.20} & 7.20 & 17.05 & 13.60 & 12.10 & 2.38 & 5.04 & 2.83 & 4.65 & 7.98 \\
int4 k-quant + ConvInt & ONNX & 0.67\,GB & CPU & 10.14 & 8.00 & 20.55 & 11.68 & 14.73 & 5.32 & 10.14 & 3.05 & 5.64 & 9.98 \\
int4 RTN & ONNX & 0.66\,GB & CPU & 8.46 & 7.30 & 18.94 & 13.52 & 12.12 & 2.42 & 5.10 & 2.85 & 4.72 & 8.05 \\
\bottomrule
\end{tabular}%
}
\end{table}

\subsection{Nemotron-0.6B --- ONNX int4 k-quant at 0.16s Delay}

\begin{table}[H]
\centering
\caption{Nemotron-0.6B int4 k-quant at 0.16\,s algorithmic delay. Config: (2,20,2), 3.2\,s history. CPU inference, batch\_size=1.}
\label{tab:app_quant_160ms}
\resizebox{\textwidth}{!}{%
\begin{tabular}{@{}llcccccccccccc@{}}
\toprule
\textbf{Variant} & \textbf{Format} & \textbf{Size} & \textbf{Device} & \textbf{Avg} & \textbf{RTFx} & \textbf{AMI} & \textbf{Earn.} & \textbf{Giga} & \textbf{LS-c} & \textbf{LS-o} & \textbf{SPGI} & \textbf{TED} & \textbf{VoxP.} \\
\midrule
int4 k-quant & ONNX & 0.67\,GB & CPU & 8.89 & 2.87 & 19.92 & 14.02 & 11.08 & 2.99 & 6.59 & 3.76 & 3.94 & 8.84 \\
\bottomrule
\end{tabular}%
}
\end{table}

\vspace{4pt}
\par\noindent{\scriptsize Earn.\,=\,Earnings22, Giga\,=\,GigaSpeech, LS-c/o\,=\,LibriSpeech Clean/Other, SPGI\,=\,SPGISpeech, TED\,=\,TED-LIUM, VoxP.\,=\,VoxPopuli. Parenthesized values are reported by original authors. A dash (--) indicates the configuration was not evaluated or results were incomplete.}

\vspace{1cm}
\hrule
\vspace{0.5cm}

\begin{center}
\textit{CoreAI, Microsoft Corporation.}\\
\textit{April 2026.}
\end{center}

\end{document}